\documentclass{article}

\usepackage[T1]{fontenc}
\usepackage{microtype}
\usepackage{graphicx}
\usepackage{booktabs}
\usepackage{multirow}
\usepackage{amsmath}
\usepackage{amssymb}
\usepackage{xspace}
\usepackage{tcolorbox}
\usepackage{xurl}
\usepackage{hyperref}
\usepackage{enumitem}

\usepackage{adaoffload-preprint}
\hypersetup{
  pdftitle={Multi-GPU MoE Inference with Adaptive Expert Offloading},
  pdfsubject={Research preprint; Creative Commons Attribution 4.0 International},
  pdfkeywords={Mixture-of-Experts, multi-GPU inference, expert offloading, CC BY 4.0}
}

\newcommand{\solution}{\textsc{AdaOffload}\xspace}

\newcommand{\postfig}{\vspace{-0.3in}}

\newcounter{finding}

\newcommand{\ziming}[1]{\textcolor[RGB]{139,69,19}{[\textbf{Ziming:} #1]}}
\newcommand{\zhongjie}[1]{\textcolor[RGB]{0,0,139}{[\textbf{Zhongjie:} #1]}}
\newcommand{\wz}[1]{\textcolor[RGB]{0,69,0}{[\textbf{William:} #1]}}
\newcommand{\yang}[1]{\textcolor[RGB]{153,153,0}{[\textbf{Yang:} #1]}}
\newcommand{\ri}[1]{\textcolor{blue}{[\textbf{Rishabh:} #1]}}
\newcommand{\ion}[1]{\textcolor{purple}{[\textbf{Ion:} #1]}}

\newif\ifshowcomments
\showcommentsfalse  %

\ifshowcomments
\else
  \renewcommand{\ziming}[1]{}
  \renewcommand{\zhongjie}[1]{}
  \renewcommand{\wz}[1]{}
  \renewcommand{\yang}[1]{}
  \renewcommand{\ri}[1]{}
  \renewcommand{\ion}[1]{}
\fi

\graphicspath{{figs/}}

\mlsystitlerunning{Multi-GPU MoE Inference with Adaptive Expert Offloading}

\begin{document}

\twocolumn[
\mlsystitle{Multi-GPU MoE Inference with Adaptive Expert Offloading}

\begin{mlsysauthorlist}
\mlsysauthor{William X. Zheng}{berkeley}
\mlsysauthor{Zhongjie Chen}{microsoft}
\mlsysauthor{Shuo Yang}{berkeley}
\mlsysauthor{Ziming Mao}{berkeley}
\mlsysauthor{Xiaoze Fan}{sjtu}
\mlsysauthor{Yi Pan}{berkeley}
\mlsysauthor{Kunjun Li}{berkeley}
\mlsysauthor{Hanfei Yu}{berkeley}
\mlsysauthor{Chenfeng Xu}{ut_austin}
\mlsysauthor{Yang Zhou}{uc_davis}
\mlsysauthor{Rishabh Iyer}{berkeley}
\mlsysauthor{Ion Stoica}{berkeley}
\end{mlsysauthorlist}
\mlsysaffiliation{berkeley}{University of California, Berkeley}
\mlsysaffiliation{microsoft}{Microsoft}
\mlsysaffiliation{sjtu}{Shanghai Jiaotong University}
\mlsysaffiliation{ut_austin}{University of Texas at Austin, Department of Computer Science}
\mlsysaffiliation{uc_davis}{University of California, Davis}

\mlsyskeywords{Machine Learning, MLSys}

\vskip 0.3in

\begin{abstract}
Modern Mixture-of-Experts (MoE) inference is constrained by GPU memory.
This constraint is mitigated by expert offloading, which stores part of the model's weight in host memory and executes on demand.
Though MoE architecture has computation sparsity, more GPUs are needed for serving increasingly larger frontier models.
However, existing expert offloading systems overlook this need and fail to scale performance with additional GPUs.
We present \solution, a general and efficient multi-GPU hybrid expert execution framework that effectively handles heterogeneous hardware and workloads.
We co-design GPU--CPU parallelism and GPU--CPU co-execution for decode and prefill with online workload observation that adapts to different hardware, workload, and batch size heterogeneity.
We evaluated \solution on five different platforms with varying GPU capability (L4, Pro6000, B200), number of GPUs, CPU capability, and NUMA topology on three different frontier open-weight models (284B to 753B).
\solution achieves $1.9\times$ to $15.5\times$ throughput speedups across different models over state-of-the-art systems.

\end{abstract}
]

\printAffiliationsAndNotice{}

\section{Introduction}
\label{sec:introduction}

Increasingly capable open-weight models make self-hosted inference attractive to organizations seeking greater control over their data and infrastructure. Frontier open-weight Mixture-of-Experts (MoE) models offer strong capabilities across a range of tasks~\citep{kimiK3Model,zaiGLM53Flash2026,deepseekV41Flash2026} and can be deployed on local hardware or rented cloud GPUs. For small organizations handling sensitive or regulated data, self-hosting offers greater control over data processing while avoiding the service quotas and recurring costs of external providers~\citep{anthropicThreatIntelligence2026,thsottiauxXServiceQuotas}.

Limited GPU memory, however, remains a key barrier to practical self-hosting. Large models commonly require multiple GPUs for their combined compute and memory capacity, and growing models and rising memory prices make fully GPU-resident deployment increasingly costly. For example, DeepSeek-V4-Flash's weights occupy approximately 145\,GiB, while its successor, DeepSeek-V4.1-Flash-1017, requires 475\,GiB---more than three times as much. KV caches and runtime buffers add to this footprint, causing memory requirements to grow further as context lengths and concurrency increase. These demands lead many users to access open-weight models only through commercial inference APIs.

The sparsity of MoE models makes expert offloading a promising way to reduce their GPU memory requirements. Since each token activates only a small subset of experts, systems can retain expert weights in host memory and cache a subset on GPUs, prefetching or fetching missing experts on demand. These transfers, however, can incur substantial overhead over bandwidth-constrained PCIe links. To reduce this overhead, recent systems employ \emph{hybrid expert execution}~\citep{kamahori2025fiddler}, executing selected experts on the CPU and transferring smaller activations and outputs instead of their weights.

Unfortunately, existing MoE offloading systems provide limited support for interactive serving at modest concurrency across multiple GPUs. Specifically, prior work has primarily focused on hybrid expert execution in single-GPU settings and for single-request inference~\citep{kamahori2025fiddler,chen2025ktransformers,yang2026freetoken,yu2026finemoe} or in exclusively throughput-oriented settings suited for large-batch offline processing~\citep{xu2026batchgen,cao2025moelightning}.

Our goal is to build a general and  efficient hybrid expert execution engine for MoE inference across diverse hardware configurations, models, and batch sizes. We target interactive services within small organizations that must meet service-level objectives (SLOs) for time to first token (TTFT) and time per output token (TPOT), typically at modest concurrency. The system should adapt to the available CPU and GPU resources and the requests being served to make effective use of each deployment's hardware.

Building a general MoE offloading engine that remains efficient across hardware configurations, models, and batch sizes requires addressing three forms of heterogeneity. The first is in how CPU and GPU resources are organized. Distributing expert computation across multiple GPUs requires balancing both the computation and the weight transfers so that each GPU and its PCIe link can contribute effectively. On the CPU, efficient execution depends on NUMA topology, memory locality, and how expert tensors are partitioned among workers. A partitioning strategy that works well for GPUs may therefore be poorly suited to CPUs, and simply assigning equal amounts of work does not ensure efficient execution. The engine must accommodate the partitioning requirements of each resource while coordinating their work to avoid transfer bottlenecks and stragglers.

The second form of heterogeneity is in the compute and transfer capabilities of different hardware platforms. The relative benefit of executing an expert on the CPU or fetching its weights for GPU execution depends on the CPU and GPU capabilities, the available PCIe bandwidth, and the amount of work assigned to that expert. These costs also interact: CPU computation and expert transfers share host memory bandwidth, while multiple GPUs may contend for the same PCIe links. Consequently, execution times measured in isolation can misrepresent the costs of concurrent CPU--GPU execution. Choosing an effective division of work requires a model that accounts for expert loads, the number of distinct experts, and contention among computation and transfers.

The third form of heterogeneity is in workload demands, which vary across models, batches, and inference phases. During decode, routing patterns and batch size determine how many experts are activated and how much work each receives, while cache residency determines which experts require fetching or CPU execution. A fixed placement or execution policy cannot accommodate these changes: executing all cache-missed experts on the CPU can leave GPU capacity idle, while fetching too many can make transfers the bottleneck. Prefill introduces another tradeoff. Longer prefills can benefit from overlapping next-layer expert prefetching with GPU computation, whereas shorter prefills can benefit from hybrid expert execution. A general engine must therefore observe the workload at runtime and adapt its execution policy as these demands change.

We present \solution, an expert-offloading engine for multi-GPU MoE inference that adapts to differences in resource organization, hardware capability, and workload demands. To accommodate CPU and GPU resource organization, \solution uses heterogeneous tensor parallelism: it shards experts across GPUs to balance computation and weight transfers, while choosing CPU partitions independently based on NUMA topology and kernel efficiency. To account for hardware capabilities, it profiles expert computation and fetching under concurrent CPU--GPU execution, capturing contention for shared resources. To adapt to workload changes, it combines these profiles with lightweight runtime observations of expert loads and cache residency to select execution policies for both decode and prefill. Put together, these mechanisms enable \solution to make effective use of available CPU and GPU resources across diverse deployments and changing workloads.

We evaluate \solution across three MoE models and five hardware platforms. The models range from 284B to 753B parameters, and the platforms span L4, PRO 6000, and B200 GPUs with varying GPU counts, CPU capabilities, and interconnect topologies. On decode-heavy reasoning workloads, \solution achieves an average of $1.9\times$--$15.5\times$ decode throughput and $54.6\%$--$85.1\%$ lower TTFT than the strongest evaluated baseline across DeepSeek-V4-Flash, GLM-5.3-Flash, and GLM-5.2. On agentic coding tasks, it improves decode throughput by an average of $4.82\times$ at modest concurrency. For DeepSeek-V4-Flash, its adaptive decode policy achieves $94.8$\% and $99.2$\% of the throughput of the best measured fixed policy at concurrency 1 and 8, respectively, approaching the performance of policies selected through an exhaustive sweep without requiring that search at runtime.

In summary, this paper makes the following contributions:
\begin{enumerate}[nosep]
\item A cost model for multi-GPU hybrid expert execution that accounts for expert loads, distinct experts, and contention between computation and transfers.
\item An inference engine, \solution, combining heterogeneous tensor parallelism with workload-adaptive CPU--GPU execution.
\item Extensive evaluation across five platforms and three MoE models, demonstrating average of $1.9\times$--$15.5\times$ decode throughput and a $54.6\%$--$85.1\%$ reduction in TTFT on reasoning workloads.
\end{enumerate}

\section{Background and Motivation}
\label{sec:background}

\subsection{Self-hosted Inference}
\label{sec:background-hardware}

\textbf{Privacy concerns and stronger open-weight models motivate self-hosted inference.} Accessing proprietary LLMs through external APIs may expose sensitive data and interaction traces to service providers, creating privacy and compliance risks. Changes in API quotas, pricing, or availability also introduce operational uncertainty. Meanwhile, open-weight models such as Kimi-K3, GLM-5.3-Flash, and DeepSeek-V4.1-Flash are narrowing the capability gap with frontier proprietary models~\citep{kimiK3Model,zaiGLM53Flash2026,deepseekV41Flash2026}. Self-hosting offers access to these capabilities while retaining control over data and infrastructure.

\textbf{Limited GPU memory remains a key barrier to self-hosted inference.} Capable open-weight models often have substantial memory requirements. For example, the weights of GLM-5.3-Flash and DeepSeek-V4.1-Flash require 176 and 475~GiB of memory under FP4 quantization, respectively, excluding runtime buffers and KV caches. For agentic and long-context workloads, KV caches can require substantial additional memory~\cite{wang2026actkv,kwon2023pagedattention}. An organization-wide inference service must handle concurrent requests while maintaining acceptable response latency and throughput. Thus, practical self-hosted inference requires fitting models within limited GPU memory while serving concurrent requests efficiently.

\begin{figure}[t]
  \centering
  \includegraphics[width=\columnwidth]{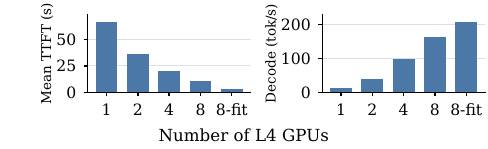}
  \postfig
  \caption{Mean TTFT and decode throughput when serving 16 concurrent requests with L4 GPUs with expert offloading (\solution). 8-fit indicates fitting all experts in VRAM.}
  \label{fig:gpu-scaling}
\end{figure}

\subsection{MoE Inference and Expert Offloading}
\label{sec:background-moe}

\textbf{MoE sparsity makes expert offloading practical.} Mixture-of-Experts (MoE) models activate only a few experts per token, a design widely adopted by recent open-weight models. This sparsity makes MoE models particularly well suited to expert offloading: expert weights can reside in host memory, with only a subset cached on GPUs. Fetch-based approaches transfer the weights of missing experts to GPUs for execution, incurring PCIe overhead when transfers cannot be overlapped with computation. Prior work uses expert prediction and prefetching to reduce cache misses~\citep{xue2024moeinfinity,song2024promoe}. Alternatively, hybrid expert execution systems execute missing experts on CPUs by transferring activations instead of expert weights~\citep{kamahori2025fiddler}. Recent work combines expert caching with bandwidth-aware scheduling to partition missing-expert execution between CPUs and GPUs~\citep{yang2026freetoken}. As Figure~\ref{fig:gpu-scaling} shows, a carefully designed expert offloading system can retain a large fraction of full-residency performance as GPU resources increase.

\begin{figure}[t]
  \centering
  \includegraphics[width=\columnwidth]{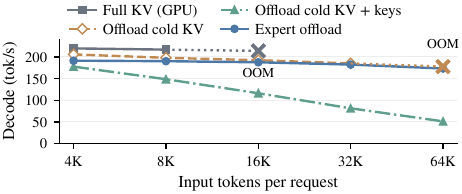}
  \postfig
    \caption{Decode throughput of DeepSeek-V4-Flash on eight L4 GPUs with 16 concurrent requests. Cold-KV offloading keeps all experts on GPU, with indexer keys on GPU or CPU as indicated. Expert offloading instead uses a half-VRAM expert cache while keeping KV and indexer keys on GPU.}
  \label{fig:kv-headroom}
\end{figure}

\textbf{Expert offloading beyond model fitting.} As models continue to scale, experts increasingly dominate the memory footprint of MoE models, accounting for over $90$\% of the parameters in many recent open-source models. Expert offloading is therefore essential when the model exceeds aggregate GPU memory. Yet its value extends beyond making the model fit. Even when all experts can reside on GPUs, keeping them resident may not be the best use of scarce VRAM. In the example with eight L4 GPUs in Figure~\ref{fig:gpu-scaling}, full expert residency leaves just $5.5$~GiB for KV caches, enough for $124$K tokens, whereas expert offloading leaves $49$~GiB, enough for $1.1$M tokens. As Figure~\ref{fig:kv-headroom} shows, this tradeoff becomes increasingly favorable for longer contexts: VRAM is better spent on repeatedly accessed KV entries than on sparsely accessed experts, as avoiding KV offloading can outweigh the overhead of fetching experts. The same tradeoff also arises under higher concurrency, as more concurrent requests further increase KV demand.
 
\begin{figure}[t]
  \centering
  \includegraphics[width=\columnwidth]{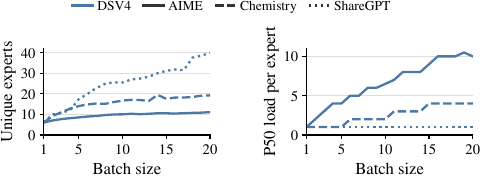}
  \postfig
  \caption{Decode expert footprint across batch sizes for DSV4-Flash
    on AIME, Chemistry~\citep{hendrycks2021mmlu}, and ShareGPT~\citep{sharegptVicuna2023} workloads.
    Left: number of unique experts; right: median token load per expert.}
  \label{fig:decode-footprint}
\end{figure}
\begin{table}[!tbp] \centering \footnotesize \setlength{\tabcolsep}{3pt} 
\caption{Comparison of \solution with prior works.}
\begin{tabular}{@{}lcccc@{}} \toprule System & \shortstack{Hardware\\aware} & \shortstack{Batch-size\\aware} & \shortstack{Multi\\GPU} & \shortstack{Supported\\parallelism} \\ \midrule KTransformers & -- & -- & $\checkmark$ & TP \\ Ollama & -- & -- & $\checkmark$ & PP \\ llama.cpp & -- & -- & $\checkmark$ & PP \\ FreeToken & $\checkmark$ & -- & -- & -- \\ \midrule \solution (ours) & $\checkmark$ & $\checkmark$ & $\checkmark$ & HTP \\ \bottomrule 
\end{tabular}  \label{tab:system-capabilities} \end{table}

\subsection{Limitations of Existing Systems}

Given the demonstrated advantages of hybrid expert execution over pure expert offloading, we focus hereafter on hybrid expert execution systems. Existing systems, such as KTransformers~\citep{chen2025ktransformers} and Ollama~\citep{ollamaFAQ2026}, statically partition experts between CPUs and GPUs during decode. This partitioning makes CPU execution the bottleneck while leaving H2D PCIe bandwidth underutilized, missing opportunities to fetch CPU-resident experts for GPU execution. These limitations worsen as the number of GPUs increases: static placement underutilizes aggregate PCIe bandwidth and exacerbates CPU--GPU imbalance, degrading performance.

FreeToken~\citep{yang2026freetoken} and Fiddler~\citep{kamahori2025fiddler} orchestrate expert execution across CPUs and GPUs to balance execution times in single-GPU settings. Extending this approach to multiple GPUs, however, is nontrivial: execution time depends on the parallelization strategy and the resulting contention for shared resources, including PCIe switch and NUMA-node bandwidth. These interactions complicate both execution-time modeling and the search for an optimal hybrid expert execution policy.

Finally, prior systems primarily target single-request inference and overlook emerging agentic workloads. Larger batches and changing workload characteristics substantially alter expert load patterns during decode, as shown in Figure~\ref{fig:decode-footprint}. Agentic workloads also exhibit variable prefill lengths that prior systems fail to accommodate. For example, some systems always favor prefetching, missing opportunities to reduce TTFT for short prefills with hybrid expert execution. Table~\ref{tab:system-capabilities} compares \solution with prior work.

Therefore, to build an efficient multi-GPU hybrid expert execution system, we must jointly consider \textbf{parallelism design} (\S\ref{design_parallelism}), \textbf{hardware heterogeneity} (\S\ref{design_offline}), and \textbf{dynamic workload patterns} (\S\ref{design_online}) to select the most efficient hybrid expert execution plan.

\section{\solution: Design}
\label{sec:design}

\subsection{Overview}

\begin{figure*}[t]
  \centering
  \includegraphics[width=\textwidth]{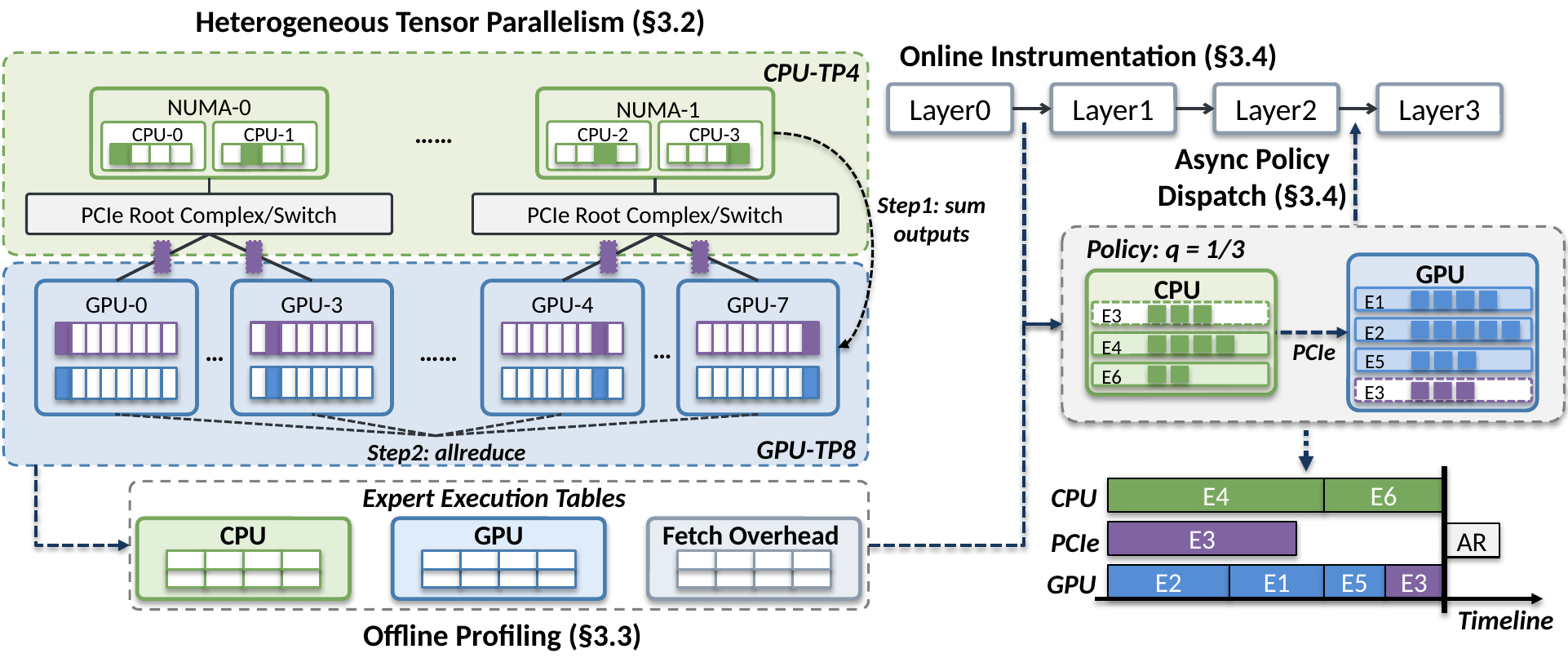}
  \postfig
  \caption{Overview of \solution. HTP employs distinct TP domains for CPUs and GPUs. The CPU is split by NUMA node; hardware profiling selects intra-NUMA sharding, resulting in TP4. Offline profiling
    provides hardware- and model-specific cost tables. Online expert-residency
    and token-load samples guide asynchronous policy updates. In this example, $q=1/3$, Expert 3 is fetched to the GPU for minimum completion time. }
  \label{fig:system-overview}
\end{figure*}

We now introduce \solution, an inference framework for hybrid expert execution that adaptively dispatches expert computation across CPUs and multiple GPUs. \solution adopts heterogeneous tensor parallelism (HTP) to efficiently coordinate CPU and GPU resources (\S\ref{design_parallelism}). HTP balances GPU workloads through tensor parallelism. For CPU execution, it selects the degree of parallelism based on NUMA topology and kernel efficiency to reduce host-memory contention and mitigate stragglers. To account for differences in hardware performance, \solution then profiles the hardware offline to accurately estimate CPU and GPU worker execution times under different expert workloads and contention conditions (\S\ref{design_offline}). Variations in workload and batch size, however, must be observed during inference. Thus, \solution combines these profiles with lightweight runtime observations of expert loads and cache residency to adapt the CPU--GPU execution policy to varying batch sizes in both the prefill and decode phases (\S\ref{design_online}). Figure~\ref{fig:system-overview} summarizes the system design.

\subsection{Heterogeneous Tensor Parallelism}
\label{design_parallelism}
Extending hybrid expert execution to multiple GPUs is not straightforward. It requires carefully balancing CPU and GPU parallelism while reconciling their differing partitioning requirements.

\textbf{Observation 1: TP better matches hybrid expert execution than PP and EP.} We consider pipeline parallelism (PP), expert parallelism (EP), and tensor parallelism (TP). PP splits layers across workers, but requires sufficient micro-batches to keep pipeline stages busy~\citep{huang2019gpipe}, making it inefficient for interactive workloads with modest concurrency. Moreover, increasing heterogeneity across model layers makes it harder to balance workloads across PP stages. EP assigns complete experts to workers and uses two all-to-all operations to dispatch and combine tokens. It is prone to load imbalance, especially with few tokens~\citep{lepikhin2021gshard}, and hybrid execution further translates routing skew into imbalanced weight transfers and CPU memory accesses: some PCIe links and NUMA nodes become overloaded while others remain underutilized. Rebalancing experts requires additional weight transfers, further stressing the bandwidth-constrained path. In contrast, TP shards each expert across all workers in a group, distributing computation and weight transfers evenly across PCIe links and NUMA nodes. In addition, the allreduce introduced by TP has a much more regular communication pattern and is more amenable to optimization over PCIe than EP's alltoall~\citep{tccl}.

\begin{figure}[!tbp]
  \centering
  \includegraphics[width=\columnwidth]{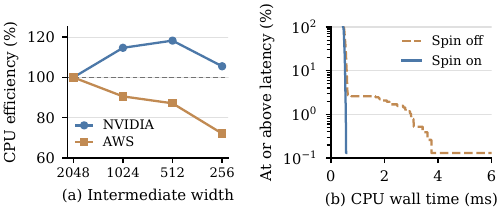}
  \postfig
  \caption{CPU execution efficiency and task latency.
    (a) CPU kernel efficiency on two different single-NUMA-node configurations at different expert intermediate widths. Intermediate widths 2048/1024/512/256 correspond to NUMA-local sharding factors of 1/2/4/8.
    (b) CPU task wall-time distributions with spin waiting off/on.}
  \label{fig:cpu-spin}
  \label{fig:expert-shard-efficiency}
\end{figure}

\textbf{Observation 2: CPU and GPU tensor-parallelism domains should be decoupled.} A naive design does not partition CPU experts and therefore suffers substantial performance degradation from NUMA memory access~\citep{chen2025ktransformers}. Applying the GPU TP degree to CPUs, or merely partitioning CPU experts by NUMA node, is also suboptimal. As shown in Figure~\ref{fig:expert-shard-efficiency}, CPU kernel efficiency within a NUMA node depends significantly on both the hardware and the intra-NUMA sharding degree.
Based on these observations, \solution adopts heterogeneous tensor parallelism, as illustrated in Figure~\ref{fig:system-overview}. HTP partitions CPU experts independently of GPU experts: it first shards host-expert tensors across NUMA nodes, then profiles CPU kernels using the model's tensor shapes to determine the optimal intra-NUMA sharding degree. The CPU TP degree, $TP_c$, is the product of the NUMA-node count and the intra-NUMA degree. In HTP, a GPU associated with each NUMA node accumulates the partial results produced by that NUMA node's CPUs, followed by a single GPU-wide allreduce.

\textbf{Mitigating CPU worker stragglers.} HTP distributes computation evenly across CPU workers, but does not guarantee equal completion times. Since the HTP reduction depends on every rank's contribution, CPU scheduling delays can expose a straggler even when ranks process equal-sized shards. We observe a long tail in CPU task completion time when CPU workers are preempted by the OS scheduler. We therefore enable spin polling for worker task queues and completion waits to reduce blocking and wake-up overhead, at the cost of additional CPU usage. Figure~\ref{fig:cpu-spin}(b) shows the reduction in CPU task tail latency.

\subsection{Profiling Compute and Transfer Costs}
\label{design_offline}

Hardware platforms exhibit substantially different characteristics, as shown in Table~\ref{tab:node-host-resources}. While HTP balances computation across GPUs and CPUs and reduces NUMA contention, \solution must also account for the capabilities of individual components and contention among them. To characterize these capabilities and interactions, \solution performs comprehensive offline system profiling.

\textbf{Observation 3: Expert loads and the number of unique experts significantly affect execution time.} For an expert $e$, matrix multiplication requires $F_e=6n_eHH_E$ FLOPs, and its weights occupy $S_e=3HH_Es_e$ bytes, where $n_e$ is the number of tokens routed to the expert, $s_e$ is the number of bytes per expert parameter, and $H$ and $H_E$ are the activation and expert hidden dimensions, respectively. We use $E_d$, $d\in\{c,g,f\}$, to denote the sets of experts executed on CPUs, executed on GPUs, and fetched from host memory, respectively. Let $L_d=\sum_{e\in E_d} n_e$ denote the expert compute load (expert-token assignments) and $U_d=|E_d|$ denote the number of unique experts. Then $T_{exec,d}$, the time required to execute the experts assigned to device $d\in\{c,g\}$, is approximated by
\begin{equation}
\label{eq:expert-exec-cost}
T_{exec, d} \approx \max(\frac{6HH_EL_d}{Comp_d},
  \frac{3HH_EU_ds_e}{MemBW_d})
\end{equation}
Similarly, the MoE execution time is approximated by
\begin{equation}
\label{eq:moe-hybrid-cost}
T_{MoE, hybrid} \approx \max(T_{exec, c},
  T_{exec, g} + \frac{3HH_EU_fs_e}{HtoDBW})
\end{equation}
On GPUs, $Comp_d$ and $MemBW_d$ remain stable, leaving $L_d$ and $U_d$ as the workload-dependent factors. On CPUs, $MemBW_d$ may also vary due to memory-bandwidth contention between expert computation and concurrent expert transfers to GPUs. Under a fixed HTP configuration, however, this contention is determined by the workload characterized by $L_d$ and $U_d$. We therefore profile expert execution as a function of these two dimensions.

\textbf{Expert execution.} For each supported decode batch size $B$, we enumerate feasible $(L,U)$ pairs, where $L$ denotes the expert-token load and $U$ the number of distinct experts. We replay a single MoE layer using the actual model shape and inference kernels, with all CPU and GPU workers running concurrently under the configured HTP setup. The resulting measurements form lookup tables $\hat{T}_{\mathrm{CPU}}(L,U)$ and $\hat{T}_{\mathrm{GPU}}(L,U)$ for estimating expert execution time.

\textbf{Fetching and resource interference.} Expert fetching and CPU execution share host-memory resources. Consequently, their costs cannot be estimated independently from isolated bandwidth measurements alone. We profile each CPU workload both without expert fetching and under sustained concurrent fetching from all GPUs. We measure fetching latency as $\hat{T}_{\mathrm{fetch}}(F\mid L_c,U_c)$, where $F$ is the number of fetched experts and $(L_c,U_c)$ describes the concurrent CPU workload. Measurements are collected with all participating GPUs fetching concurrently to capture the actual contention under HTP. These measurements provide CPU execution costs with and without concurrent fetching, as well as fetching costs conditioned on the CPU workload.

\begin{figure}[!tbp]
  \centering
  \includegraphics[width=\columnwidth]{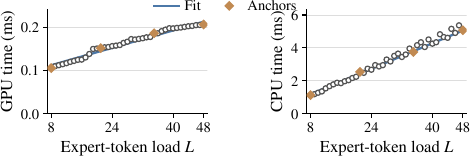}
  \postfig
  \caption{Four-point latency prediction on L4$\times$8 for DSV4-Flash
    at $U=8$. Held-out WAPE is 2.6\% and 4.0\%, respectively.}
  \label{fig:linear-regression-relax}
\end{figure}

\textbf{Heuristics to reduce profiling space.} A batch of $b$ tokens with top-$k$ expert routing contains $bk$ expert-token routes. The expert-token load $L$ and unique expert count $U$ satisfy $1\le L\le bk$ and $\left\lceil\frac{L}{b}\right\rceil\le U\le\min(|E|,L)$, where $|E|$ is the number of routed experts per layer. For DSV4-Flash with $b=32$, $k=6$, and $|E|=256$, this yields 18,048 feasible $(L,U)$ pairs. This space is prohibitively expensive to profile exhaustively. When the number of unique experts is fixed, computation time scales linearly with the expert-token load $L$. For batch sizes greater than 8, we use linear regression for extrapolation. As shown in Figure~\ref{fig:linear-regression-relax}, four-point linear regression captures the relationship between execution latency and expert-token load $L$. Across the configurations that we exhaustively profiled, the average WAPE of the four-point linear regression is 1.99\%.

\subsection{Online Adaptation to Workload Changes}
\label{design_online}

Figure~\ref{fig:decode-footprint} shows how expert footprints vary across decode batch sizes and workloads. For instance, at a batch size of 20, the ShareGPT workload activates 3.6 times as many unique experts as AIME, while its median token load per expert is significantly lower. However, workloads and batch sizes change during inference and cannot be fully characterized offline. Therefore, \solution uses low-overhead online instrumentation to capture the expert footprint at runtime.
Instrumented forward passes write statistics into preallocated device buffers, which are copied asynchronously to host buffers for policy optimization. For each sampled forward pass, we record expert and cache statistics, including per-layer unique experts, expert load, cache-missed experts, and the fetching order determined by the fetching policy (e.g., most recently used first or highest load first). Experiments on DSV4 show that instrumentation slows instrumented forward passes by 1.3--6.6\%. These statistics provide the expert load $L$ and unique expert count $U$ for the current workload.
The sampling interval and the number of historical samples retained for policy selection are configurable. A shorter sampling interval allows the policy to adapt more quickly to workload changes, at the cost of greater instrumentation overhead. Retaining fewer historical samples makes policy selection more responsive to recent changes, but potentially less stable and more sensitive to noise.

\textbf{Asynchronous Policy Selection and Execution.}
\label{sec:design-policy-select}
The hybrid execution policy is determined by the fraction of cache-missed experts fetched to GPUs, with the objective of minimizing $T_{MoE, hybrid}$. We denote this fraction by $q\in[0,1]$. Let $Q$ denote the candidate set of $q$ values considered by the policy selection algorithm. In our implementation, \solution uses $Q=\{0,1/64,2/64,\ldots,62/64,63/64,1\}$. Policy selection runs asynchronously without blocking the main inference path. The main inference path adopts updated policies at forward-pass boundaries.

\paragraph{Decode}
The policy selector uses the recorded forward-pass statistics to evaluate candidate $q$ values. Formally,
\begin{equation}
a^* = \arg\min_{a \in \mathcal{A}}
\sum_{\ell=1}^{N_{\mathrm{layers}}}
 \hat{T}_{\ell}(\mathcal{S}|a),
\mathcal{A} = \{\mathrm{hybrid}(q) : q \in Q\}
\label{eq:policy-selection}
\end{equation}

Here $\hat{T}_{\ell}(\mathcal{S}|a)$ represents the estimated execution time of layer $\ell$ given the sampled workload $\mathcal{S}$ and the execution policy $a$. It is obtained by reconstructing the expert assignment induced by $a$ from the sampled statistics. For a candidate $q$, we determine the number of experts to fetch and select them according to the fetching priority order. This yields the expert sets $E_g$ and $E_c$ assigned to each device, and the instrumentation provides their expert loads and unique expert counts. These quantities index the offline profiles $\hat{T}_{\mathrm{CPU}}(L_c,U_c)$, $\hat{T}_{\mathrm{GPU}}(L_g,U_g)$, and $\hat{T}_{\mathrm{fetch}}(F\mid L_c,U_c)$. The CPU estimate accounts for contention during concurrent fetching. We substitute these component estimates into Eq.~\eqref{eq:moe-hybrid-cost} to obtain $\hat{T}_{\ell}(\mathcal{S}|a)$. When multiple historical samples are available, we choose the policy $a$ that minimizes the average estimated execution time across the samples.

\paragraph{Prefill.}
An alternative to on-demand hybrid execution is to prefetch all expert weights for the next layer, so weight transfer overlaps with computation for the current layer. This suits prefill, where the many routed tokens activate nearly all experts. Prior work instead prefetches only the experts predicted to be activated~\citep{zhong2025hybrimoe}. However, the performance of these approaches varies substantially with prediction accuracy, and prediction complicates integration with CUDA Graphs. Thus, we consider prefetching all experts in the next layer. The execution time of an MoE layer is then
\begin{equation}
\label{eq:moe-prefetch-cost}
T_{MoE, prefetch} = \max(T_{exec, g}, \frac{\sum_{e \in E}S_e}{HtoDBW})
\end{equation}
The prefill phase uses the same policy-selection formulation, with $\mathcal{A}=\{\mathrm{prefetch}\}\cup\{\mathrm{hybrid}(q):q\in Q\}$.

\section{Evaluation}
\label{sec:evaluation}
\begin{table*}[t]
  \caption{Hardware specifications across different multi-GPU nodes. Effective compute is measured under one representative workload.}
  \label{tab:node-host-resources}
  \centering
  \small
  \setlength{\tabcolsep}{4pt}
  \renewcommand{\arraystretch}{1.15}
  \begin{tabular}{@{}lcclcccc@{}}
    \toprule
    Provider & GPUs & \shortstack{Eff. compute\\(TFLOP/s)} & CPU (cores) & \shortstack{Eff. compute\\(TFLOP/s)} & DRAM (GiB) & \shortstack{PCIe/link\\(GB/s)} & \shortstack{Aggregate H2D\\(GB/s)} \\
    \midrule
    Mibura & L4 $\times 4$       & 3.390 & Xeon 4410Y (24)    & 0.428 & 256 & Gen4 (32) & 98.0 \\
    GCP    & L4 $\times 8$       & 6.317 & Xeon (48)          & 0.510 & 402 & Gen3 (16) & 33.3 \\
    NVIDIA & PRO 6000 $\times 2$ & 6.829 & TR PRO 7955WX (16) & 0.314 & 480 & Gen5 (64) & 103.2 \\
    AWS    & PRO 6000 $\times 2$ & 8.751 & Xeon 8559C (24)    & 0.463 & 500 & Gen5 (64) & 50.8 \\
    \bottomrule
  \end{tabular}
\end{table*}

\subsection{Experimental Setup}
\textbf{Testbed.} We conduct experiments on four different machines as listed in Table~\ref{tab:node-host-resources}.
The GCP and AWS machines are rented from the corresponding cloud service providers.
Other machines are workstations that are accessed through research partnerships.
We conduct experiments on them without modifying any of their hardware configurations.
These machines exhibit substantial heterogeneity in PCIe bandwidth and CPU compute capacity.
For example, NVIDIA Pro6000 has GPUs in different PCIe switch resulting in higher PCIe aggregate bandwidth while 
AWS Pro6000 has both GPUs sharing the same PCIe switch. This allows us to evaluate how effectively \solution accommodates diverse hardware characteristics.

\textbf{Baselines.} We compare \solution against llama.cpp~\citep{llamacppServing2026} (b76b133), KTransformers~\citep{chen2025ktransformers} (5632f71), and Ollama~\citep{ollamaFAQ2026} (d919272),
three SOTA hybrid expert execution systems that support multi-GPUs. KTransformers integrates the SGLang~\citep{zheng2024sglang} backend to support TP, while llama.cpp and Ollama both use PP.
We use their default multi-GPU supports and keep the model checkpoints and quantization identical.  
For a fair comparison, all systems use the same VRAM expert cache size.

\textbf{Model.} All systems serve DeepSeek-V4-Flash with MXFP4 quantization, 
with half of the available GPU memory reserved for the expert cache. 
We primarily focus on DeepSeek-V4-Flash because it's the largest model that can fit into all machines' DRAM while all baseline systems have native support on.
We also evaluate GLM-5.3-Flash with NVFP4 quantization and leave the result figures in Appendix to show the generalizability of our results across different model architectures.

\textbf{Workloads.} 
We evaluate the end-to-end performance of all systems on two representative LLM workloads: (1) mathematical reasoning and (2) agentic coding. For reasoning, we use AIME24~\citep{aime2024Dataset} and AIME25~\citep{aime2025Dataset}, two challenging mathematical reasoning benchmarks.
For agentic coding, we use SWE-bench Verified~\citep{jimenez2024swebench,swebenchVerifiedDataset} with Claude Code~\citep{anthropicClaudeCode2026} as the agent harness.
AIME is a single-turn, decode-heavy workload with 206 (495) median (p99) input sequence length (ISL) and 4032 (23902) median (p99) output sequence length (OSL).
In contrast, agentic coding is a multi-turn workload with long initial input to describe the task and multiple rounds of tool call and response.
For example, for our evaluation of \solution at concurrency of 8, the initial task description has median of 9339 tokens. 
The subsequent tool call results have median of 332 tokens and median OSL per turn is 155.
We randomly sample a set of problems from each benchmark and issue them in the same order to all systems to ensure a consistent workload. 
We use 1, 4, and 8 as concurrency levels.

\textbf{Metrics.} We primarily focus on decode throughput (per-request average token per second) and time-to-first-token (TTFT) (per-request average time to receive the first token).

\subsection{End-to-end Result on Realistic Workloads}
\begin{figure*}[t]
    \centering
    \includegraphics[width=\textwidth]{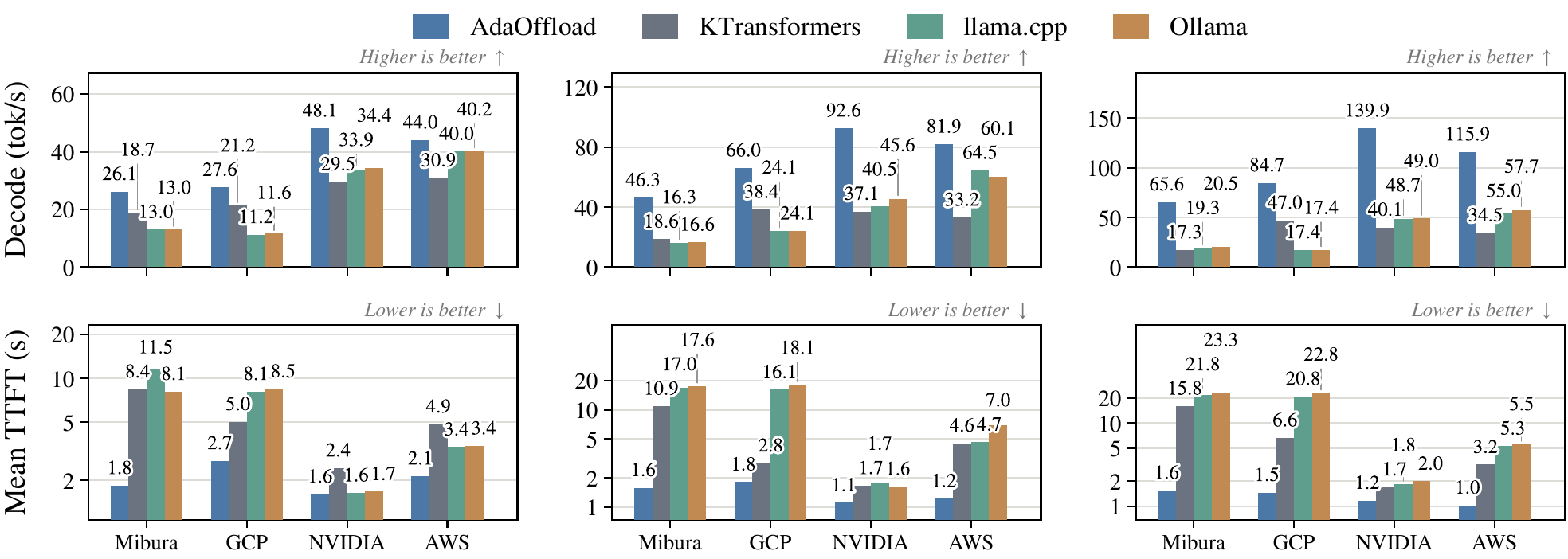}
    \postfig
    \caption{End-to-end performance of \solution and baselines on AIME reasoning tasks. Columns show concurrency 1, 4, and 8; rows show streaming decode throughput and mean TTFT.}
    \label{fig:e2e-aime}
\end{figure*}

\begin{figure*}[t]
    \centering
    \includegraphics[width=\textwidth]{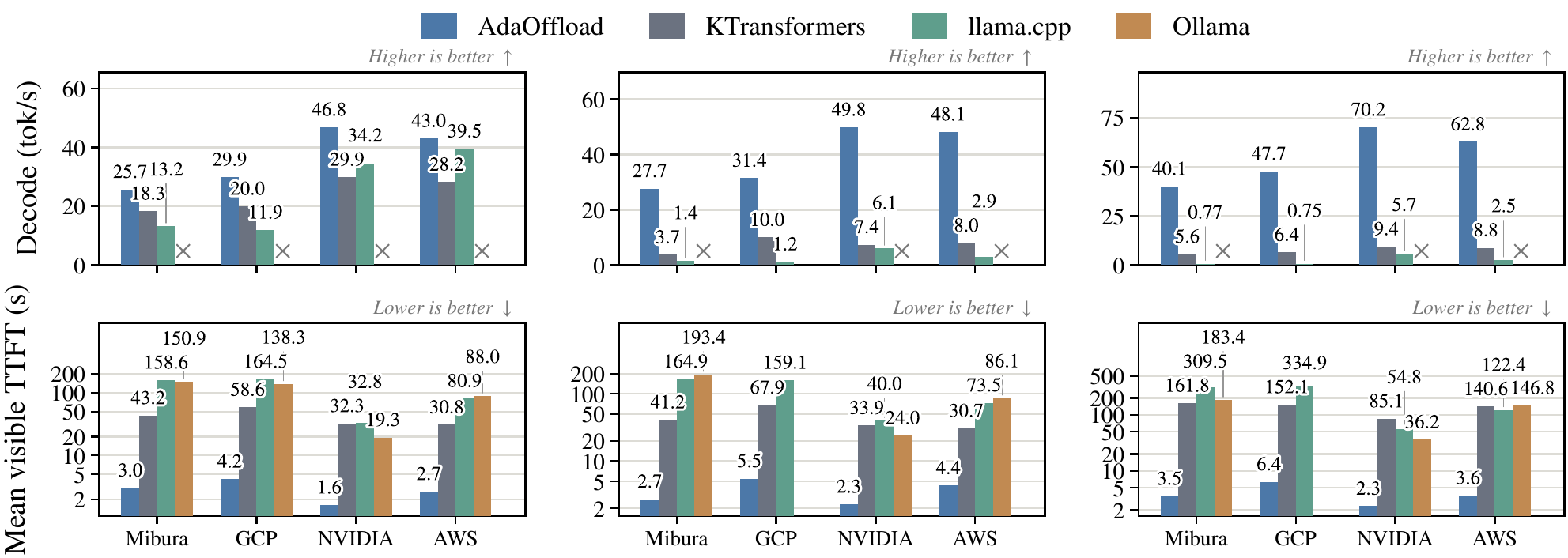}
    \postfig
    \caption{End-to-end performance of \solution and baselines on coding tasks. Columns show concurrency 1, 4, and 8; rows show streaming decode throughput and mean visible TTFT. Ollama timeouts all session except for concurrency 1.}
    \label{fig:e2e-claude-code}
\end{figure*}

\textbf{Mathematical reasoning.}
As shown in Figure~\ref{fig:e2e-aime}, \solution achieves an average improvement across four different platforms in decode throughput (TTFT reduction) of 1.30$\times$ (41\%), 1.88$\times$ (57\%), and 2.47$\times$ (67\%) over the strongest baseline at three different concurrency levels.
\solution outperforms the baselines at small batch size due to hybrid execution.
At larger batch sizes, \solution outperforms the baselines more because it adaptively selects the hybrid execution policy.

\textbf{Agentic coding.}
Figure~\ref{fig:e2e-claude-code} shows the end-to-end performance of \solution and baselines on coding tasks.
As concurrency increases, all baseline systems suffer from prefix cache misses and poor prefill efficiency,
leading to low session decode throughput overall.
For almost all cases, other frameworks reach the 30-minute session timeout limit. 
In contrast, \solution consistently performs efficient prefill via adapting between hybrid execution and full-layer prefetching.
This leads to the substantial gap in overall decode throughput for the baseline systems.

\subsection{Effectiveness of Adaptive Hybrid Execution}
\label{sec:effectiveness-hybrid}
We next examine whether \solution adjusts hybrid expert execution policy efficiently under varying workloads and hardware configurations. We study its policy decisions separately for decode and prefill, respectively.

\begin{figure}[t]
    \centering
    \includegraphics[width=\columnwidth]{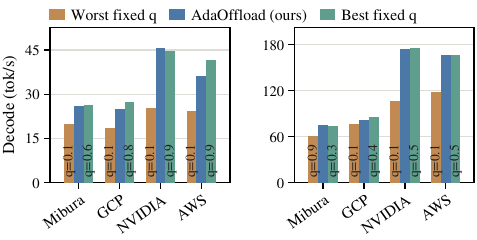}
    \postfig
    \caption{Decode throughput of \solution versus the best and worst measured fixed-$q$ policies at concurrency $C=1$ (left) and $C=8$ (right).}
    \label{fig:q-sensitivity}
\end{figure}

\textbf{Decode.} 
We first evaluate the accuracy of \solution's adaptive policy in selecting the fraction $q$ of experts to fetch to GPUs. Figure~\ref{fig:q-sensitivity} compares \solution's selection with the best and worst fixed-$q$ policies identified through brute-force search. Averaged across the four platforms, \solution attains 94.8\% and 99.2\% of the throughput achieved by the best fixed-$q$ policy at $C=1$ and $C=8$, respectively, demonstrating that it accurately identifies near-optimal execution policies without exhaustive search. Meanwhile, \solution outperforms the worst fixed-$q$ policy by 49.1\% and 33.9\%. 

The optimal $q$ varies substantially across hardware configurations and concurrency levels, making a fixed policy ineffective. In most cases, $q=0.1$ performs the worst because it fetches too few experts, underutilizing PCIe bandwidth and GPU compute. By contrast, the Mibura testbed (L4$\times$4) favors a smaller $q$ under high concurrency. Its smaller expert cache incurs 2--3$\times$ more expert cache misses per forward pass, making expert fetching more expensive than CPU execution. These contrasting cases demonstrate that \solution accurately adapts its policy to different hardware configurations and runtime conditions.
 
\begin{figure}[t]
    \centering
    \includegraphics[width=\columnwidth]{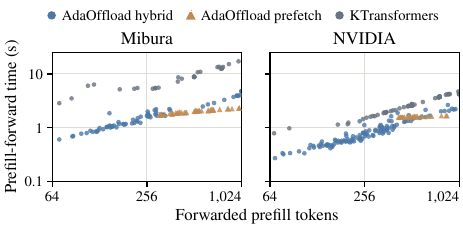}
    \postfig
    \caption{Prefill forward latency of \solution and KTransformers for coding tasks at concurrency $C=8$.}
    \label{fig:prefill-hybrid}
\end{figure}

\textbf{Prefill.} 
For prefill, \solution dynamically switches between CPU co-execution and next-layer prefetch based on the prefill length. 
Figure~\ref{fig:prefill-hybrid} compares its prefill forward latency with KTransformers on the coding workload from our end-to-end evaluation. 
We omit Ollama and llama.cpp, as frequent prefix cache misses force them to repeatedly prefill long contexts, resulting in poor performance. 
On the two platforms, \solution switches to next-layer prefetch at 300 and 400 tokens, respectively, based on the runtime observation and estimated computation and PCIe transfer costs. 
On Mibura, the hybrid execution policy chooses q-value of around 0.6 to leverage the CPU compute capacity when prefill length is between 128 and 300.
On NVIDIA, given its powerful PCIe bandwidth, the hybrid execution policy chooses q-value of 1 to fully utilize the GPU resources.
This adaptive switching keeps prefill latency low across varying prefill lengths.

\subsection{Scalability of Hybrid Expert Execution}

We next evaluate how \solution scales hybrid expert execution across multiple GPUs with HTP (\S\ref{design_parallelism}). We vary the number of GPUs and compare \solution against KTransformers and an offload-only variant of \solution that disables hybrid expert execution. We run eight concurrent reasoning tasks from the AIME dataset and measure decode throughput and mean TTFT. For all configurations, both KTransformers and \solution are allowed to utilize all available CPU resources.

\begin{figure}[!tbp]
    \centering
    \includegraphics[width=\columnwidth]{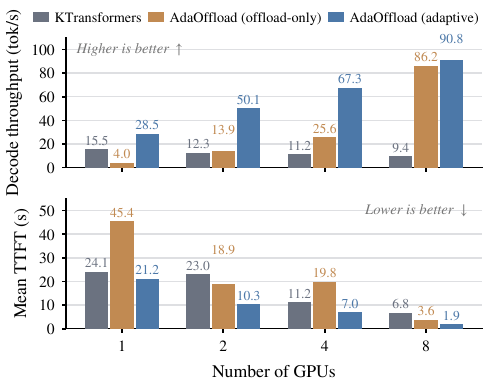}
    \postfig
    \caption{Decode throughput and mean TTFT for different numbers of GCP L4 GPUs. Framework can use all CPUs.}
    \label{fig:eval-scalability}
\end{figure}

\textbf{Decode.} 
As shown in Figure~\ref{fig:eval-scalability}(top), KTransformers fails to translate additional GPUs into higher decode throughput, which even decreases as the GPU count increases. This is because its fixed CPU--GPU expert placement during decoding cannot adapt to additional GPU capacity, leaving the added GPU resources underutilized. In contrast, \solution's offload-only variant executes all experts on GPUs by fetching miss expert on demand and scales from 4 tok/s with one GPU to 86.2 tok/s with eight GPUs, but leaves CPU compute resources underutilized. 
With HTP, \solution jointly exploits CPUs and GPUs while optimizing their parallelism independently, scaling from 28.5 to 90.8 tok/s. In summary, \solution achieves 1.84$\times$--9.61$\times$ higher decode throughput than KTransformers across 1--8 GPUs.

\textbf{Prefill.} 
As shown in Figure~\ref{fig:eval-scalability}(bottom), KTransformers scales with additional GPUs, reducing TTFT from 24.1\,s to 6.8\,s during prefill. However, its fixed hybrid CPU--GPU execution is suboptimal for most cases, where GPU prefetch is more efficient. \solution instead adapts to the prefill length, using GPU prefetch in most cases and hybrid expert execution only when beneficial. For the latter, HTP and the adaptive execution policy efficiently distribute expert execution across CPUs and GPUs. With 1, 2, 4, and 8 GPUs, 
\solution achieves 1.14$\times$, 2.24$\times$, 1.61$\times$, and 3.67$\times$ TTFT speedup over KTransformers, respectively.
The adaptive configurations use 32 CPU expert workers: one cross-NUMA pool with
interleaved expert weights at one GPU, and two NUMA-local 16-worker pools at
2--8 GPUs. KTransformers uses 48 CPU expert workers throughout.

\subsection{System Overhead}
When performing instrumentation, \solution shows an average of $4.04\%$ extra latency on sampled forwards.
\solution captures CUDA graph for supported decode batch size. On average, the CUDA graph consumes $0.47$ GiB of GPU memory when support $4$ different batch sizes $(1,2,4,8)$ in our evaluation.
The asynchronous decode policy solver takes an average of $4.1$ ms to decide the policy for a batch size.

\subsection{Performance on Other MoE Models and GPUs}
We further evaluate \solution on GLM-5.3-Flash across the same four platforms and GLM-5.2 on B200 to test if \solution could generalize to other models and newer platforms. (See appendix for detailed results.)
GLM-5.3-Flash and GLM-5.2 have 320B and 753B parameters respectively and each token is routed to top-8 experts while DeepSeek-V4-Flash routes to top-6 experts.

For GLM-5.3-Flash, Averaged across the four platforms, \solution achieves 3.86$\times$, 7.64$\times$, and 9.78$\times$ decode throughput over the strongest baseline at concurrency 1, 4, and 8, respectively, while reducing mean TTFT by 79.6\%, 70.0\%, and 83.9\%.
Comparing with the best fixed-$q$ policy, \solution attains 94.7\% and 99\% of the best measured throughput.
GLM-5.2 is tested with two NVIDIA B200 GPUs with half of the HBM reserved for expert cache (Result figure in Appendix).
On this two-GPU configuration, \solution achieves $6.42\times$, $17.98\times$, and $22.17\times$ decode throughput over the strongest baseline at concurrency 1, 4, and 8, respectively, while reducing mean TTFT by $84.8\%$, $84.4\%$, and $86.0\%$.

\section{Related Work}
\label{sec:related-work}

\paragraph{Expert caching and prefetching.}
Expert offloading uses GPU memory as a cache for a larger expert pool.
Mixtral-offloading combines expert caching with speculative
prefetching~\citep{eliseev2023offloading}, while MoE-Infinity uses
request-level activation traces to guide cache replacement and
prefetching~\citep{xue2024moeinfinity}.
ProMoE predicts subsequent expert use from intermediate
results~\citep{song2024promoe}; FineMoE additionally exploits fine-grained
expert-selection patterns and prompt semantics~\citep{yu2026finemoe}.
LayerScope coordinates prediction and cross-layer scheduling to account for
prefetch cost and PCIe bandwidth competition~\citep{yu2026layerscope}.
These techniques reduce demand misses or move transfers off the critical
path, but prefetched weights still consume device memory and link bandwidth.
A separate direction reduces the work or bytes required for a miss:
HOBBIT loads less critical experts at lower precision~\citep{tang2024hobbit},
and SPICE combines speculative prefetching with approximate expert
surrogates and exact CPU recovery~\citep{lyu2026spice}.
These approximation choices are distinct from deciding where to execute a
given expert computation.

\paragraph{CPU--GPU collaborative execution.}
Host-resident experts need not always be transferred to a GPU.
Fiddler executes selected expert work on the CPU to avoid expensive weight
movement~\citep{kamahori2025fiddler}.
KTransformers develops optimized CPU expert kernels and heterogeneous
CPU/GPU execution~\citep{chen2025ktransformers}, while HybriMoE combines
dynamic intra-layer scheduling with inter-layer prefetching and cache
management~\citep{zhong2025hybrimoe}.
CoX-MoE coalesces expert computation across a batch and coordinates
AMX-enabled CPU execution, GPU-resident experts, and attention
offloading~\citep{son2026coxmoe}.
We build on FreeToken, which combines an elastic expert cache with a
measured-bandwidth model that divides residual expert
work between CPU execution and GPU loading, within a graph-compatible
serving runtime~\citep{yang2026freetoken}.
Its evaluation focuses on single-GPU deployments. We extend this execution
setting to a node in which multiple GPUs share host resources and
also communicate with one another. The CPU/GPU split must therefore be
considered together with expert placement and the selected parallelism.

\paragraph{Multi-GPU offloading and pipelining.}
Batching can amortize weight movement across more tokens.
FlexGen optimizes tensor placement and execution for latency-insensitive
batched inference~\citep{sheng2023flexgen}.
MoE-Lightning pipelines CPU, GPU, and I/O work using paged weights and a
hierarchical roofline model; it also serves large MoE models on multiple
low-cost T4 GPUs~\citep{cao2025moelightning}.
Klotski schedules multiple batches around expert activation and loading
requirements to reduce pipeline idle time~\citep{fang2025klotski}.
These systems focus on large batch offline processing, while we primarily focus on interactive 
serving with medium batch sizes.
In our case, batch size will not be large enough to cover the overhead of pipelining.

\section{Conclusion}
\label{sec:conclusion}
We present \solution, an expert-offloading engine for multi-GPU MoE inference that adapts to differences in resource organization, hardware capability, and workload demands.
We co-design GPU--CPU parallelism and GPU--CPU co-execution for decode and prefill with online workload observation that adapts to different hardware, workload, and batch size heterogeneity.
Across a range of heterogeneous GPU platforms ranging from low-power GPU to high-end GPU, \solution achieves $1.9\times$--$15.5\times$ throughput speedup over state-of-the-art systems across three different frontier MoE models, DeepSeek-V4-Flash, GLM-5.3-Flash, and GLM-5.2.

\bibliographystyle{mlsys2025}
\bibliography{references}

\section*{AI Usage Disclosure}

We used OpenAI Codex as the generative-AI assistant across several stages of this work.
It assisted with the implementation and debugging guided by human design and review.
For manuscript, it assisted generating experiment result figures, proofreading, and formatting.

\clearpage
\appendix
\twocolumn[{
\begin{minipage}{\textwidth}
  \centering
  \toptitlebar
  {\Large\bfseries\scshape
    Multi-GPU MoE Inference with Adaptive Expert Offloading\par
    (Appendix)\par}
  \bottomtitlebar
\begin{table}[H]
  \caption{Model specification.}
  \label{tab:model-memory}
  \centering
  \small
  \setlength{\tabcolsep}{4pt}
  \renewcommand{\arraystretch}{1.15}
  \begin{tabular*}{\textwidth}{@{\extracolsep{\fill}}lcccccc@{}}
    \toprule
    Model &
    \shortstack{Expert\\format} &
    \shortstack{Total params\\(B)} &
    \shortstack{Active params\\(B)} &
    \shortstack{Expert weights\\(GiB)} &
    \shortstack{Non-expert\\weights (GiB)} &
    \shortstack{Total weights\\(GiB)} \\
    \midrule
    DSV4-Flash     & MXFP4 & 284 & 13 & 137.06 &  8.24 & 145.30 \\
    GLM-5.3-Flash  & NVFP4 & 320 & 18 & 159.47 & 16.62 & 176.09 \\
    GLM-5.2        & NVFP4 & 753 & 40 & 379.69 & 34.65 & 414.34 \\
    \bottomrule
  \end{tabular*}
\end{table}

\begin{figure}[H]
  \centering
  \includegraphics[width=\textwidth]{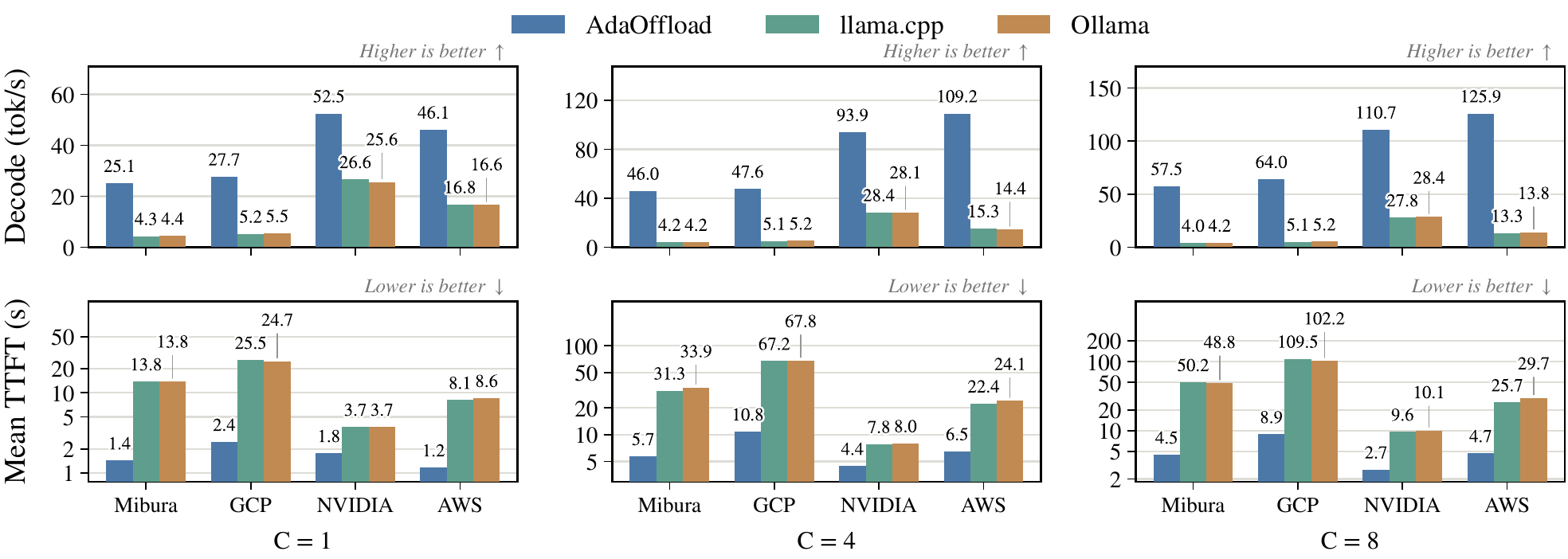}
  \caption{End-to-end performance of \solution and baseline on AIME reasoning tasks with GLM5.3-Flash NVFP4.}
  \label{fig:glm53-aime}
\end{figure}
\end{minipage}
}]
\thispagestyle{empty}

\section{Model Specifications}
\label{app:hardware-model-specifications}

Detailed model specification for DeepSeek-V4-Flash, GLM-5.3-Flash, and GLM-5.2 we evaluated. These two families of models are representative frontier open-weight models with different architecture.
GLM routes each token to 8 experts while DeepSeek model routes each token to 6 experts.
For all models, routed experts occupy more than 90\% of the total model weights.

\section{GLM-5.3-Flash AIME Evaluation}
\label{app:glm53-aime}

Figure~\ref{fig:glm53-aime} extends the AIME evaluation to
GLM-5.3-Flash with NVFP4 routed experts, comparing \solution with
llama.cpp and Ollama on the four platforms.
KTransformers does not have official support for GLM-5.3-Flash FP4 so we omit it in the comparison.

\begin{figure}[!htbp]
  \centering
  \includegraphics[width=\columnwidth]{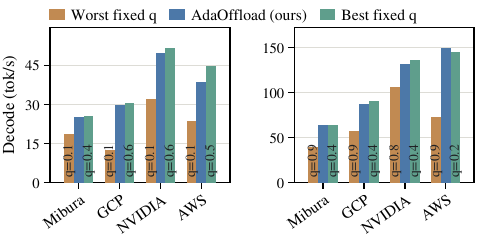}
  \caption{Decode throughput of \solution versus the best and worst measured fixed-q policies at concurrency C=1 and C=8 for GLM5.3-Flash.}
  \label{fig:glm53-q-sensitivity}
\end{figure}

\paragraph{Decode policy sensitivity.}
Figure~\ref{fig:glm53-q-sensitivity} compares \solution's adaptive decode
policy with the best and worst measured fixed-$q$ policies for GLM-5.3-Flash.
We sweep $q=0.1,0.2,\ldots,0.9$ and then run the adaptive policy, using one
persistent server per platform with matched priming and cache resets.

For GLM-5.3-Flash, Averaged across the four platforms, \solution achieves 3.86$\times$, 7.64$\times$, and 9.78$\times$ decode throughput over the strongest baseline at concurrency 1, 4, and 8, respectively, while reducing mean TTFT by 79.6\%, 70.0\%, and 83.9\%.
Comparing with the best fixed-$q$ policy, \solution attains 94.7\% and 99\% of the best measured throughput.

\begin{figure}[!t]
  \centering
  \includegraphics[width=\columnwidth]{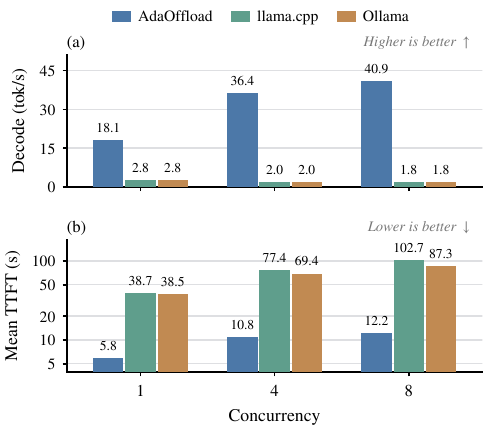}
  \caption{AIME performance of \solution, llama.cpp, and Ollama with
    GLM-5.2 NVFP4 on two B200 GPUs as concurrency varies.
    (a) Aggregate streaming decode throughput; (b) mean TTFT (log scale).
    The output cap is 512 tokens. TTFT includes the first request.}
  \label{fig:glm52-b200-aime}
\end{figure}

\newpage
\null\newpage
\section{GLM-5.2 AIME Evaluation on B200}
\label{app:glm52-b200-aime}

Figure~\ref{fig:glm52-b200-aime} compares \solution with llama.cpp and
Ollama when serving GLM-5.2 with NVFP4 routed experts on two
NVIDIA B200 GPUs. 
KTransformers does not natively support GLM-5.2 and is therefore omitted from comparison.
The configuration reserves half of the available HBM
for the expert cache. 
We use 2, 8, and 16 AIME questions at concurrency
1, 4, and 8, respectively, with scheduled arrivals every 3 seconds.

With two B200, the aggregate GPU HBM capacity is 360\,GiB, smaller than the routed expert size for GLM-5.2.
This example illustrates an important role \solution plays in inference ecosystem.
Renting B200 is very expensive. 
For people who want to serve frontier model like GLM-5.2, \solution gives the option of avoiding renting 4 B200 to fully fit the model's weight.
This provides a budget option for small-concurrency inference of strong model on a small number of powerful GPUs.

\end{document}